\documentclass[10pt,twocolumn,letterpaper]{article}

\usepackage{cvpr}
\usepackage{times}
\usepackage{epsfig}
\usepackage{graphicx}
\usepackage{amsmath}
\usepackage{amssymb}

\usepackage{subfigure}
\usepackage{multirow}
\usepackage{algorithm}
\usepackage{algorithmic}
\usepackage{bm}
 \usepackage{color}

\usepackage[pagebackref=true,breaklinks=true,letterpaper=true,colorlinks,bookmarks=false]{hyperref}

\cvprfinalcopy 

\def\cvprPaperID{1483} 

\ifcvprfinal\fi
\begin{document}

\title{Studying Very Low Resolution Recognition Using Deep Networks}

\title{Studying Very Low Resolution Recognition Using Deep Networks}

\author{Zhangyang Wang, Shiyu Chang, Yingzhen Yang, Ding Liu, and Thomas S. Huang\thanks{Zhangyang Wang and Thomas Huang's research works are supported in part by US Army Research Office grant W911NF-15-1-0317.}
 \\
Beckman Institute, University of Illinois at Urbana-Champaign, Urbana, IL 61801, USA\\
{\tt\small \{zwang119, chang87, yyang58, dingliu2, t-huang1\}@illinois.edu}
}

\maketitle

\begin{abstract}

Visual recognition research often assumes a sufficient resolution of the region of interest (ROI). That is usually violated in practice, inspiring us to explore the Very Low Resolution Recognition (VLRR) problem. Typically, the ROI in a VLRR problem can be smaller than $16 \times 16$ pixels, and is challenging to be recognized even by human experts. We attempt to solve the VLRR problem using deep learning methods. Taking advantage of techniques primarily in super resolution, domain adaptation and robust regression, we formulate a dedicated deep learning method and demonstrate how these techniques are incorporated step by step. Any extra complexity, when introduced, is fully justified by both analysis and simulation results. The resulting \textit{Robust Partially Coupled Networks} achieves feature enhancement and recognition simultaneously. It allows for both the flexibility to combat the LR-HR domain mismatch, and the robustness to outliers. Finally, the effectiveness of the proposed models is evaluated on three different VLRR tasks, including face identification, digit recognition and font recognition, all of which obtain very impressive performances.
\end{abstract}

\section{Introduction}

While object recognition research has witnessed substantial achievements so far, it is often assumed that the region of interest (ROI) is large enough and contains sufficient information for recognition. However, this assumption usually does not hold in practice. One typical example is face recognition from video surveillance \cite{VFR}. Due to the prohibitive costs of installing high-definition cameras all around, most surveillance systems have to rely on cameras of very limited definitions. Besides, wide-angle cameras are normally used in a way that the viewing area is maximized. In turn, the face region in the scene can be extremely small and low-quality. In a text recognition system \cite{text}, cheap and versatile cameras make it possible to quickly scan documents, but their low definitions also present challenges for robust character segmentation and recognition. While similar problems are ubiquitous for recognition in the wild, a principled approach to solve them is highly desirable.




Regrettably, the general \textit{Very Low Resolution Recognition} \textbf{ (VLRR)} problem has been largely overlooked, except for certain existing efforts in face recognition \cite{VFR, synthesis}.  Empirical studies \cite{lui2009meta} in face recognition proved that a minimum face resolution between $32 \times 32$ and $64 \times 64$ is required for stand-alone recognition algorithms. An even lower resolution results in a much degraded recognition performance \cite{lui2009meta, cvpr08} for conventional recognition models. The severe information loss from HR to LR makes it unlikely to extract or recover sufficient recognizable features directly from LR subjects \cite{VFR}. Typically, the ROI in a VLRR problem can be smaller than $16 \times 16$ pixels, and is even difficult (but still possible) to be recognized by human viewers.

In this paper, we make the first attempt to solve the VLRR problem using deep learning methods \cite{Alex}. Starting from the simplest baseline, we perform a step-by-step model evolution, gradually obtaining more sophisticated and powerful models. Any extra complexity, when introduced, is fully justified by both analysis and simulation results. The final outcome, named \textit{Robust Partially Coupled Networks}, achieves feature enhancement and recognition simultaneously. It is equipped with both the flexibility to combat the cross-resolution domain mismatch, and the robustness to outliers. The proposed models are applied to resolving real VLRR problems on three different tasks, including face identification, digit recognition and font recognition, all of which obtain very impressive performances.

\subsection{Problem Definition}
In real-world settings, VLRR directly recognizes visual subjects from low-resolution (\textbf{LR}) images, without any pre-defined high-resolution (\textbf{HR}) image. Here, we introduce HR images as ``auxiliary variables'' to our model training, by assuming each training image has both LR and HR versions available. As verified by our following experiments, HR images help discover more discriminative features which are prone to be overlooked from LR images. In testing, there are only LR images available, and the VLRR model is applied without any need of HR images.

There is indeed no absolute boundary between LR and HR images. However, the literature in object and scene recognition have observed a significant performance drop, when the image resolution is decreased below 32 $\times$ 32 pixels (see Fig. 2 of \cite{80tiny}). In \cite{VFR}, the authors also reported a degradation of the recognition performance, when face regions became smaller than 16 $\times$ 16 pixels. 

We follow \cite{80tiny, VFR} to choose the LR image resolution to be no more than $16 \times 16$, and HR no less than $32 \times 32$\footnote{Note that our developed models can apparently be applied, even when the standard of LR changes.}. In this paper, we treat the original training images as HR images, unless otherwise specified. To generate LR images for training (and also testing), we downsample the original images by a factor of $s$, and then upscale them back to the original resolution, by nearest neighbor (NN) interpolation. The upscaling operation is intended to ensure sufficiently large spatial supports for hierarchal convolutions, as well as to facilitate feature transfer. Since NN interpolation does not bring in any new information, \textit{the upscaled images are treated as our default LR images hereinafter}.

In summary, the problem is defined as: \textit{learning a VLRR model from training images which have both LR and HR versions, and applying the model on LR testing images}.

\section{Model I: Basic Single Network}
\subsection{Motivation}
Deep convolutional neural networks (CNNs) \cite{Alex} have recently shown an explosive popularity, partially due to its prevailing success in image recognition tasks, on various subjects such as faces \cite{sun2014deep}, digits \cite{SVH}, texts \cite{ECCVtext} and fonts \cite{ICLR}. However, all the models assume reasonable resolutions of ROIs \cite{Alex}. Popular datasets, such as LFW \cite{LFW} and ImageNet \cite{Alex}, mostly come with moderate to high image resolutions (typically around several hundred pixels per dimension). It remains unexplored whether the performances of conventional CNNs are still reasonable for VLRR tasks. We will thus start from investigating the basic single CNN for VLRR, as the \textbf{simplest baseline}.

\subsection{Technical Approach}

\begin{figure}[bpht]
\centering
\begin{minipage}{0.49\textwidth}
\centering {
\includegraphics[width=\textwidth]{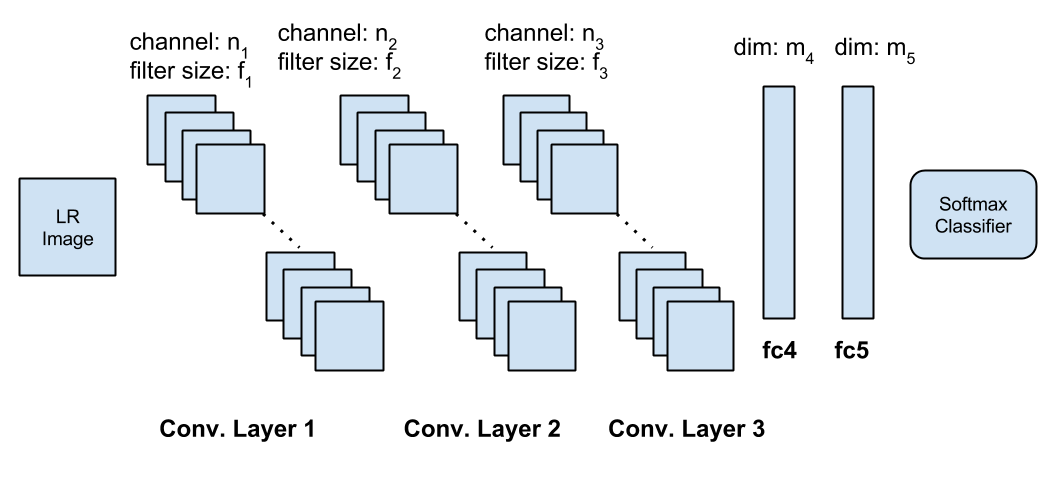}
}\end{minipage}
\caption{The basic architecture of conventional convolutional neural network (CNN). }
\label{fig:CNN}
\end{figure}
Fig. \ref{fig:CNN} illustrates the our Model I (basic CNN) that is similar to the popular ImageNet structure \cite{Alex}, with 3 convolutional layers and 2 fully-connected layers. Since LR images in VLRR problems do not have too much (low- and high-level) information to be extracted hierarchically, we do not refer to deeper architectures. For the $i$-th ($i$ = 1, 2, 3) convolutional layer, we assume it of $n_i$ channels, filter sizes of $f_i$, and the stride of 1. Note that a LR image (less than $16 \times 16$) in VLRR problems contains little self-similarity to be utilized, and is often corrupted. Therefore as verified in our experiments, applying large convolutional filter sizes to extract patches (as in \cite{Tang}, etc.) brings little benefits. The two fully-connected layers have the dimensionality of $m_4$ and $m_5$, respectively. The network is trained \textbf{from end to end}, as suggested by \cite{Alex} and many others.


\subsection{Simulation}

We adopt the popular CIFAR-10 and CIFAR-100 datasets \cite{cifar}, as our simulation subjects. The CIFAR-10 dataset consists of 60,000 32 $\times$ 32 color images, in 10 classes, with 6000 images per class. For each class, there are 5,000 images for training and 1,000 for testing. The CIFAR-100 dataset is just like the CIFAR-10 and of the same total volume, except it has 100 classes containing 600 images each. In all experiments, we convert images to gray scale for model simplicity. The original (HR) images are first downscaled by $s$ = 4 into 8 $\times$ 8. They are then upscaled back to 32 $\times$ 32 by NN interpolation, as the LR images. For each LR image, we subtract its mean and normalize its magnitude, which are later put back to the recovered HR image. Small additive Gaussian noise ($\sigma$ = 0.05) is added as the default data augmentation in training.
%

We implement our models using the cuda-convnet package \cite{Alex}. ReLU is adopted as the neuron. The batch size is 128. The learning rate starts from 0.1 and is divided by 10 when the training curve reaches a plateau. Dropout is also applied. We also compare the VLRR models trained on LR inputs with those trained on original HR images. The size of the last fully-connected layer is always chosen to be the number of classes, e.g., $m_5$ = 10 for CIFAR-10 and 100 for CIFAR-100. $m_4$ is fixed to be 1024 by default. We then vary other network configurations to examine how the baseline performances are affected, which are outlined in Table \ref{config}. The following important insights are readily concluded:
\begin{itemize}
\item The performances of VLRR models with LR input images are largely degraded compared to those obtained on HR images.
\item In VLRR experiments, larger filter sizes are hardly beneficial, because VLRR subject cannot afford as much spatial resolution (and detail) loss as HR images.
\item While in HR experiments, adding more filter channels generally helps, it may hurt VLRR performances. That is likely to be the result of model overfitting, since LR subjects contain scarce visual features.
\end{itemize}
In addition, increasing the size of $m_4$ marginally rises the performance, while bringing in significant complexity. We also tried to repeat Conv. Layer 3 for deeper architectures, ending up with the observation that increasing depth does not contribute visibly to VLRR, since LR images do not have rich visual semantic information. The above findings demonstrate the distinct characteristics of VLRR, that \textbf{VLRR problems do not benefit as much from larger filter sizes, more filter channels or deeper models, as conventional visual recognition problems do}. Therefore, we adopt the following moderate configuration as default: $n_1$ =  64, $f_1$ =  5; $n_2$ = 64, $f_2$ = 3; $n_3$ = 32, $f_3$ = 1; $m_4$ = 1024.

 \begin{table}[tbp]
\footnotesize
 \begin{center}
 \caption{The top-1 classification error rates (\%) of various network configurations, with either LR or HR training images, on the CIFAR-10 and CIFAR-100 datasets.}
 \vspace{0.1em}
 \label{config}
 \begin{tabular}{|c|c|c|c|c|c|c|c|c|c|}
 \hline
$n_1$ & $n_2$ & $n_3$ & $f_1$ & $f_2$ & $f_3$ &  \multicolumn{2}{|c|}{CIFAR-10} & \multicolumn{2}{|c|}{CIFAR-100} \\\cline{7-10}
 &  & & & & &  LR & HR & LR & HR \\
 \hline
 5 & 3 & 1 & 64 & 64 & 32  &  \textbf{27.47} & 14.41 & \textbf{45.03} & 38.37 \\
 \hline
 7 & 3 & 1 & 64 & 64 & 32  &  28.86 & 14.24 &  49.82 &  37.07 \\
 \hline
 5 & 5 & 1 & 64 & 64 & 32  &  28.82 & 14.06 & 50.01 &  36.69 \\
 \hline
 5 & 3 & 3 & 64 & 64 & 32  &  29.13 & \textbf{13.78} & 49.95 & \textbf{35.02}  \\
 \hline
 5 & 3 & 1 & 32 & 32 & 32  &  30.03 &17.91 & 50.81 &  42.23 \\
 \hline
 5 & 3 & 1 & 64 & 32 & 32  &  27.87 & 15.45 &  48.91 & 40.27  \\
 \hline
  5 & 3 & 1 & 64 & 64 & 64  &  28.08 & 14.32 & 46.67 &  37.61 \\
 \hline
 \end{tabular}
 \end{center}
 \end{table}

\section{Model II: Single Network with Super-Resolution Pre-training}

\subsection{Motivation}

Table \ref{config} reminds that directly classifying LR visual objects are unreliable and prone to overfitting, since their visual features are scarce and highly degraded. On the other hand, it  is noteworthy that although HR images are not available in real testing, it could still be utilized in training as auxiliary information to obtained enhanced features.

Classical paradigms first apply a super-resolution (SR) algorithm \cite{SR} to a LR image and then classify its SR result \cite{VFR}. Recently, the SR performance has been noticeably improved, with the aid of deep network models \cite{Zhaowen, DJSR}. However, the recovered HR images still suffer from not only inevitable oversmoothness and detail loss, but also various artifacts introduced by the reconstruction process \cite{baker2002limits}, that further undermine the subsequent recognition performance. The authors in \cite{VFR} incorporated discriminative constraints to learn the relationship between the HR and LR face images for recognition. In \cite{cvpr08}, class-specific facial features were included in SR as a prior. Such ``close the loop'' approaches perform consistently better than traditional two-stage (SR then recognition) pipelines.


\subsection{Technical Approach}
\begin{figure}[bpht]
\centering
\begin{minipage}{0.49\textwidth}
\centering {
\includegraphics[width=\textwidth]{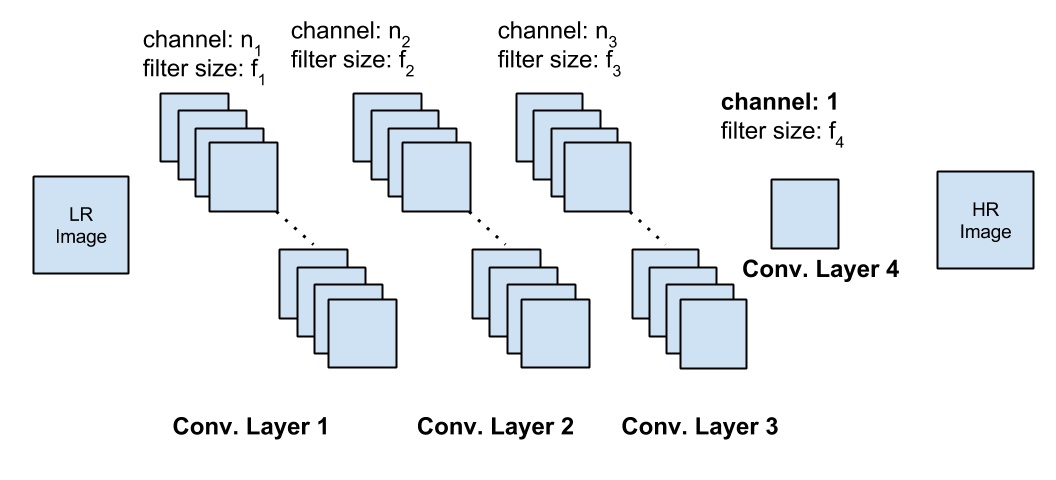}
}\end{minipage}
\caption{The super resolution (SR) sub-network, as the pre-training part of Model II. Note it does not contain the fully-connected recognition sub-network, which is the same as in Fig. 1 (fc4, fc5 and softmax). The new Conv. 4 Layer has only one channel, and is discarded after pre-training.}
\label{SR_VLRR}
\end{figure}

Model II is a ``decomposed'' version of Model I into two sub-networks: the super resolution (SR) sub-network for unsupervised pre-training, and the recognition sub-network for supervised fine-tuning. As outlined in Fig. \ref{SR_VLRR}, the SR sub-network consists of four convolutional layers, that takes LR images as inputs and HR as outputs. It is trained in an unsupervised way to predict the nonlinear mapping between LR and HR images. Compared to the convolutional parts in Fig. \ref{fig:CNN}, it is easy to notice the newly added Conv. 4 Layer, that has only one channel. It has been revealed in \cite{Tang} that such a layer acts as a linear filter that averages its input feature maps to produce the final reconstruction. After the SR sub-network is trained, the Conv. 4 Layer is \textbf{DISCARDED}, and the two fully-connected layers as well as a softmax classifier are added on its top. The entire network, which now has exactly the same topology as Model I, are jointly tuned in a supervised manner.

Earlier multi-layer feature learning methods, such as the Auto-Encoder and the Denoising Auto-Encoder, perform a pre-training of weights using unlabeled data alone for reconstruction, followed by supervised fine-tuning \cite{semi3}. The previous work \cite{erhan2009difficulty} showed that unsupervised pre-training appeared to play a regularization role predominantly. The Model II shares the similar idea and targets the unsupervised part at SR, instead of reconstruction. The final supervised fine-tuning correlate those unsupervised features with class labels. The resulting network aims to achieve resolution enhancement and recognition simultaneously.


\subsection{Simulation}

We follow the default network configuration and the same experiment setting as in Section 2.2. The LR and HR images are first utilized to train a SR subnetwork, with a upscaling factor of 4. CIFAR-10 witnesses a drop of 2.32\% in the classification error rate (ending up to be 25.15\%), while on CIFAR-100 the error rate is reduced to 46.50\%, with a decrease of 3.45\%.

\noindent \textbf{Remark:} Since SR is a one-to-many problem, any one of the ``many'' is a plausible solution, and it is impossible to ensure that the added details are authentic to the original HR image. In contrast, object recognition tries to identify the one from ``many''. Then, one may have a bit concern in the reasoning why the SR pre-training could possibly improve the visual recognition performance? Our explanation is, while not guaranteed to be faithful, the hallucinated details from SR pre-training help discover subtle but discriminative features, that benefit recognition. They are otherwise prone to be overlooked from LR images.

\section{Model III: Single Network with Pre-training and LR-HR Feature Transfer}

\subsection{Motivation}


Though limited improvements are obtained by introducing SR pre-training, a large performance gap of around 10\% on CIFAR-10/100 still exists, between models trained with LR and HR inputs. Why is there such a performance gap? One hypothesis is that the SR pre-training process does not bring in enough discriminative ability for recognition purposes \cite{lin2004fundamental}. The hypothesis is further evidenced, when we look at the visualizations of filters learned from either HR or LR inputs. While there are clearly visible edges and curves in most HR filters, LR filters learn much fewer recognizable patterns, and suffer from more ``dead'' filters. Our idea is to discover more discriminative features, by blending HR images in the training data.



\noindent \textbf{A Domain Adaptation Viewpoint}. Our goal is equivalent to learning more \textbf{transferable} features from HR to LR, which resembles the domain adaption scenarios. HR images constitute the source domain, and LR images are the target domain. It was hypothesized \cite{Bengio09} that the intermediate representations could be shared across variants of the input distribution. The model in \cite{sentiment} learned a cross-domain feature extraction, using unlabeled data from all domains. A classifier is trained on the transformed labeled data from the source domain, and tested on the target domain.

\noindent \textbf{A Data Augmentation Viewpoint}. Since in Model II, HR images were already in the training set, it may look questionable that why HR inputs may still help SR, if no new information would be added. Our explanation is: blending HR images with the original LR training data could be treated as a problem-specific data augmentation approach for SR pre-training. Common data augmentations would corrupt or modify the training images in many ways. Similarly, the introduction of HR images is alternatively viewed, as to augment the original LR images, by transforming them into HR versions. The difference is that it \textbf{enhances} the training data, rather than degrading them.

\subsection{Technical Approach}

\begin{figure}[bpht]
\centering
\begin{minipage}{0.49\textwidth}
\centering {
\includegraphics[width=\textwidth]{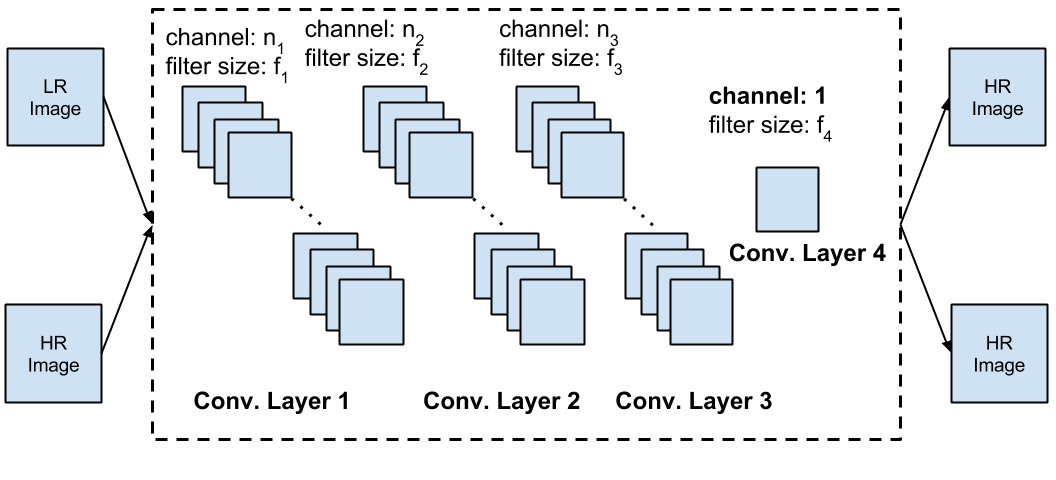}
}\end{minipage}
\caption{The super resolution (SR) sub-network enhanced by LR-HR feature transfer, as the pre-training part of Model III. Note it does not contain the fully-connected recognition sub-network, which is the same as in Fig. 1 (fc4, fc5 and softmax).}
\label{FCN}
\end{figure}

The super resolution (SR) sub-network of Model III, as shown in Fig. \ref{FCN}, is related to the domain adaption techniques from \cite{Bengio09, sentiment, coupledAE}. In the context of domain adaption, we treat LR and HR images to be from two related but non-overlapping domains. The SR sub-network could be viewed as as two \textbf{fully-coupled} channels. One LR-HR channel reconstructs the HR domain samples from the LR domain samples, which is essentially equivalent to the SR sub-network of Model II. The other HR-HR channel reconstructs the HR domain samples from themselves, who is supposed to learn more discriminative features than the former one. By enforcing the two channels to use fully-shared hidden layers, the model aims to learn cross-domain, transferable features. With the shared filters from the HR-HR channel, the LR-HR channel is also expected to obtain enhanced features for the SR pre-training task.




\subsection{Simulation}
We first pre-train the unsupervised SR model enhanced by feature transfer in Fig. \ref{FCN}, followed by supervised tuning. All experiment settings follow previous ones, except that we find when training such a ``hybrid'' model, choosing smaller learning rates helps a faster and more steady convergence. In practice, we start with the learning rate of 0.01 and do not anneal it during training. On CIFAR-10, the error rate is reduced to 21.72\%, while on CIFAR-100 the error rate goes down to 43.03\%.


\section{Model IV: Partially Coupled Networks}

\subsection{Motivation}

From Model I to Model III, we have obtained up to 10\% accuracy gains in our CIFAR-10/100 simulation experiments. Is there any further room for improvements?

We take a step back and re-think about the key prerequisite in Model III: we utilize both LR and HR data to learn a fully shared feature representation across domains. Since each LR-HR pair indicates the same scene or object, it is reasonable to assume that there exists a hidden space that are \textbf{approximately} shared by LR and HR features \cite{lin2005coupled, yang2010image}. However, the assumption that the LR and HR data should have \textbf{strictly} equal feature representations in the coupled subspace, appears to be overly strong.

As explored previously \cite{wang2012semi}, the mappings between LR and HR spaces are complex, spatial-variant and nonlinear. Taking face recognition for example, \cite{lui2009meta} empirically showed that the similarity measures in the LR and HR spaces were not identical, so that the direct matching of a LR probe image within the HR gallery performed poorly. The domain mismatch between LR and HR images is further enlarged, due to the noise, blur, and other corruptions when generating LR images. 



We make a bold hypothesis that \textbf{LR and HR image features do not fully overlap even after highly sophisticated transforms}. Therefore, we attempt to ``relax'' Model III a bit, in order for certain flexibility to cope with feature variances, due to the mismatches of LR and HR domains.

\subsection{Technical Approach}

\begin{figure}[htbp]
\centering
\begin{minipage}{0.47\textwidth}
\centering {
\includegraphics[width=\textwidth]{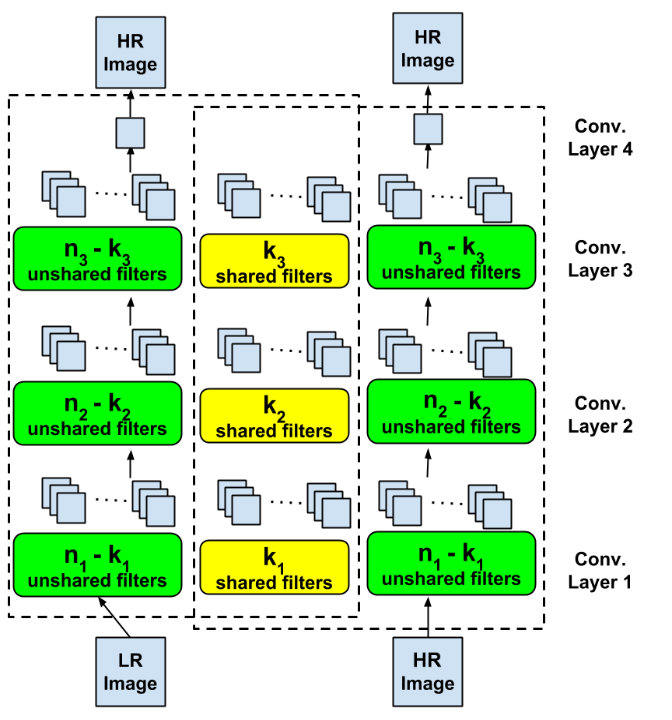}
}\end{minipage}
\caption{The Partially Coupled SR Networks (PCSRN), as the pre-training part of Model IV. Note it does not contain the fully-connected recognition sub-networks. For each channel, its recognition sub-network is the same as in Fig. 1 (fc4, fc5 and softmax). The two channels share $k_i$ convolutional filters in the $i$-th layer.}
\label{PCN}
\end{figure}

Fig. \ref{PCN} depicts \textit{Partially Coupled SR Networks} (PCSRN), as the pre-training part of Model IV. PCSRN is built with two unsupervised channels in Fig. \ref{SR_VLRR}. The major difference between PCSRN and the SR subnetwork in Model III (Fig. \ref{FCN}), lies in that the former are ``loosely'' coupled by only sharing $k_i$ convolutional filters in each layer ($i$ = 1, 2, 3). They tend to capture the commonality feature patterns across both LR and HR domains \footnote{These shared features are usually the basic correspondences that mostly persists under resolution changes, such as pixel intensity histogram, strong edges, and specific structural layouts \cite{lui2009meta}, that make up the foundation for cross-resolution recognition}. Based on our above hypothesis, PCSRN also allows each single channel to learn domain-specific features by $m_i - k_i$ unshared filters ($i$ = 1, 2, 3), that complement the shared features and handle the mismatches. Fig. \ref{FCN} could also be viewed as a special case of  Fig. \ref{PCN} when $m_i = k_i$.

After PCSRN is pre-trained, we follow the same routine to replace Conv. Layer 4 by two fully-connected layers, followed by a softmax classifier independently on each top. Each channel (including the shared filters) has exactly the same topology as Model II or III. The whole two-channel Model IV is fine-tuned in a supervised manner. Either channel classifies its own input, while the two parts still interact by taking advantage of shared filters. In testing, we ``decouple'' the model and use only the left LR-HR channel in Fig. \ref{PCN}, including all the shared filters.

The ``partially coupled'' architecture manifests our important hypothesis, that the domain mismatches in VLRR problems are only reduced but do not vanish even after learned transforms. It gives rise to important extra flexibility to transfer knowledge across domains. Despite looking like a reckless idea, its benefits will be validated next.

\subsection{Simulation}

Fig. \ref{PCN} outlines the general idea, but it is still unclear whether and how much performance gains could be obtained from the more complicated architecture. In addition, how many of the filters should be coupled? To answer them, we conduct a quantitative research on the CIFAR-10/100 cases. We fix the numbers of filters $n_1$, $n_2$ and $n_3$, and vary the numbers of shared filters $k_1$, $k_2$ and $k_3$. We define the \textit{``coupled ratio''} for layer $i$: $c_i$ = $k_i$/$n_i$  ($i$ = 1, 2, 3). Note when $c_i = 0$ ($i$ = 1, 2, 3), the model splits into two uncorrected parts and is equivalent to Model II. On the contrary, $c_i = 1$  ($i$ = 1, 2, 3) leads to the fully-coupled Model III.

 \begin{table}[tbp]
\footnotesize
 \begin{center}
 \caption{The top-1 classification error rates (\%) with different configurations of coupled filters, on the CIFAR-10 and CIFAR-100 datasets.($n_1$ = 64, $n_2$ = 64, $n_3$ = 32)}
 \vspace{0.2em}
 \label{search}
 \begin{tabular}{|c|c|c||c|c|c||c|c|}
 \hline
$k_1$ & $k_2$  & $k_3$& $c_1$ & $c_2$ & $c_3$ & CIFAR-10 & CIFAR-100  \\
 \hline
 0 & 0 & 0 &  0 & 0 & 0 & 25.15 & 46.50   \\
 \hline
  0 & 0 & 16 &  0 & 0 & 0.50 &   21.81 & 43.48  \\
 \hline
  0 & 0 & 32 &  0 & 0 & 1.00 &   22.85 & 44.03  \\
 \hline
  0 & 32 & 16 &  0 & 0.50 & 0.50 & 21.43 &  43.27 \\
 \hline
  0 & 64 & 16 &  0 & 1.00 & 0.50 & 21.87 &  44.09 \\
 \hline
  32 & 32 & 16 &  0.50 & 0.50 & 0.50 & 21.04 & 43.21   \\
 \hline
  32 & 32 & 24 &  0.50 & 0.50 & 0.75 & 19.91 & \textbf{41.52}    \\
 \hline
  32 & 32 & 32 &  0.50 & 0.50 & 1.00 & 20.72 & 41.90    \\
 \hline
  32 & 48 & 24 &  0.50 & 0.75 & 0.75 & \textbf{18.77} &  41.64    \\
 \hline
  32 & 64 & 24 &  0.50 & 1.00 & 0.75 & 19.52 &   42.99    \\
 \hline
 48 & 48 & 24 &  0.75 & 0.75 & 0.75 & 19.06 & 42.15   \\
 \hline
  64 & 64 & 32 & 1.00 & 1.00 & 1.00 & 21.72  & 43.03  \\
 \hline
 \end{tabular}
 \end{center}
 \end{table}

To identify the best configuration,  we perform a course grid search on each $c_i$ ($i$ = 1, 2, 3), from 0 to 1 with a step size of 0.25. Even so, a ``brutal-force'' search leads to training models with at least $5^3$ = 125 configurations. To avoid such an exhaustive effort, we develop a strategy by referring to the common belief in domain adaptation \cite{sentiment}, that the discrepancy between two domains is expected to be gradually reduced with layers goes deeper \cite{coupledAE}. Hence we always give priority to increasing $c_i$ of deeper layers (starting from Conv. Layer 3, then 2 and 1). Once the error rate tends to rise, we roll back the last change, and instead increase the coupled ratio of its immediate lower layer. As summarized in Table \ref{search}, the results strongly    support to our previous ``non-overlapping'' hypothesis. In fact, all partially coupled models that we tried in Table \ref{search} (0  \textless  $c_i$ \textless 1) obtain better performances than Model II ($c_i$ = 0), while most of them are superior to Model III ($c_i$ = 1) as well. Moreover, it is observed that higher layers are more sensitive to over-regularization (i.e., being overly coupled ) than lower layers. For example, simply increasing $c_3$ from 0.75 to 1.00, when $c_1$ and $c_2$ are both fixed at 05, raise the error rate on CIFAT-10 from 19.91\% to 20.72\%. As a result, we choose $c_1$ = 0.50, $c_2$ = 0.75 and $c_3$ = 0.75, as our default configuration hereinafter. The resulting model gradually narrows the gap between the two domains with increasing number of layers, but still allows for certain flexibility of each channel for domain-specific features.

We have verified that the partially coupled architecture could lead to additional gains compared to the fully-coupled model. However, it remains unresolved how to adaptively choose the best $c_i$ for specific cases, instead of ad-hoc trials. One potential solution is to learn a mapping function that regularizes the intrinsic relationship between the two domain-specific representations, such as in \cite{wang2012semi}. Beyond VLRR, the ``partially coupled''  architecture could be potentially applied to many other cross-domain recognition problems. We leave it as future work.

\section{Model V: Robust Partially Coupled Networks}

\subsection{Motivation}

So far, with the help of the SR pre-training as well as the flexible ``partially-coupled'' domain adaptation, we obtain satisfactory results on the CIFAR-10/100 LR data, which are artificially downsampled from clear HR groundtruth. However, typical real data from low-resolution sources, such as video surveillance, are usually accompanied with sensor noise and impulsive outliers. Even in high-resolution cases, there are outlier factors that may destroy the manifold structure badly. One such example could be found in face recognition: since faces are neither perfectly convex nor Lambertian \cite{lui2009meta}, face images taken under directional illumination often suffer from self-shadowing, specularity, or saturations in brightness. The pose and expression changes also introduce more spatially localized artifacts, that makes a fraction of the data grossly corrupted.

We adopt Mean-Square Error (MSE) as the loss function in SR pre-training, who could be rather sensitive to outliers. One may argue that MSE works widely well in most unsupervised deep learning models, e.g., Burger et al. \cite{plain} showed that the plain multi-layer perceptrons can produce decent results and handle different types of noise. In fact, it has been observed that the learning ability of deep systems could be weakened by these outliers, which are pervasive in real-world data \cite{chen2015efficient}. A few robust learning methods have been proposed to immunize the harmful influences caused by outliers. A correntropy-induced loss function is proposed to combat outliers \cite{qi2014robust}. The authors of \cite{xu2014deep} concatenated another specific denoising CNN to handle the complex and signal-dependent outliers, at the cost of the growing parameter volume due to the merge.

 \begin{table}[htbp]
 \begin{center}
 \footnotesize
 \caption{The top-1 classification error rates (\%), with MSE or huber loss functions in SR pre-training, on the clear and corrupted CIFAR-10/100 datasets.}
 \vspace{0.2em}
 \label{table_huber}
 \begin{tabular}{|c|c|c|c|}
 \hline
& &  CIFAR-10 & CIFAR-100 \\
 \hline
 $\multirow{2}{*}{0\% (clean)}$ & MSE (Model IV) &  18.77 & 41.64 \\
 \cline{2-4}
$$ & huber (Model V) &  18.81 & 41.70 \\
 \hline
$\multirow{2}{*}{5\% corrupted}$ & MSE (Model IV) & 19.99 & 43.43 \\
 \cline{2-4}
$$ & huber (Model V)& 19.04 & 42.18 \\
 \hline
 $\multirow{2}{*}{10\% corrupted}$ & MSE (Model IV) & 23.03 & 46.20 \\
 \cline{2-4}
$$ & huber (Model V)& 20.28 & 43.79 \\
 \hline
 $\multirow{2}{*}{15\% corrupted}$ & MSE (Model IV) & 25.05 & 49.49 \\
 \cline{2-4}
$$ & huber (Model V)& 22.24 & 46.18 \\
 \hline
 \end{tabular}
 \end{center}
 \end{table}

 \begin{table*}[htbp]
 \begin{center}
\footnotesize
 \caption{Review and Comparison of All Models. ($*$: the error gains are in the presence of 15\% outliers.)}
 \vspace{0.2em}
 \label{compare}
 \begin{tabular}{|c|c|c|c|c|}
 \hline
& \textbf{Improvement Over} & \textbf{Add} &  \multicolumn{2}{|c|}{\textbf{Error Rate Reduction} (\%)} \\
\cline{4-5}
 &  \textbf{The Last Model} & \textbf{Complexity?} & CIFAR-10 & CIFAR-100 \\
  \hline
Model II & exploit HR training data and perform SR pre-training & No & 2.32 & 3.45\\
 \hline
 Model III & learn transferable feature from HR data to enhance the SR pre-training & No & 3.43 & 3.47\\
 \hline
  Model IV & introduce extra flexibility and avoid over-regularization by the  partially-coupled structure& Yes & 2.95 & 4.86\\
 \hline
  Model V & adopt the huber loss in pre-training to be robust to outliers & No & 2.81$^*$ & 3.31$^*$\\
 \hline
 \end{tabular}
 \end{center}
 \end{table*}

\subsection{Technical Approach}

We replace the MSE loss of Model IV, with the convex and continuous \textbf{huber loss} function \cite{huber}, in SR pre-training. We name it \textit{Robust Partially Coupled Networks}, as our Model V. The huber loss is widely used in robust regression for its lower sensitivity to outliers. It comprises two parts, corresponding to the $\ell_2$ and $\ell_1$ losses \cite{bach2011convex}, formally defined as:
\begin{equation} \label{huber}
H_c(x, y)  = \left \{ \begin{array}{ll}
\frac{1}{2}(x - y)^2 &  \quad \textup{if}\quad |x - y| < c  \\
c|x - y| - \frac{c^2}{2} &  \quad \textup{if}\quad  |x - y| \ge c\\
\end{array} \right.
\end{equation}
The cutting edge parameter $c$ intuitively specifies the maximum value of error, after which the $\ell_1$ penalty has to be applied. We are aware that more robust losses could be designed,  and choose huber loss here because it is easy to implement and calculate.

In VLRR, we do find that the outliers possess a much stronger negative impact (we will verify it in simulation) than in conventional recognition, since LR images may not supply a sufficiently representative feature distribution to sketch the correct data manifold. The use of huber loss greatly alleviates the problem.

\subsection{Simulation}

We adopt the ``clean'' CIFAR-10/100 data, while creating ``corrupted'' versions by adding salt-and-pepper impulse noise on 5\%, 10\% and 20\% of randomly selected pixels before downsampling. We then test Model VI and V on both datasets, whose only difference lies in the choice of MSE or huber loss in SR pre-training. Dropout is applied to both models. We fix the empirical value $c$ = 1.345 in (\ref{huber}) as suggested by \cite{huber} and many others. During pre-training, we observe that huber loss appears to accelerate the model convergence a bit. As in Table \ref{table_huber}, the performances of huber loss is almost identical (marginally worse) than MSE in the outlier-free cases, but becomes superior in the presence of outliers, and finally gains a remarkable margin of up to 3\% over MSE with 15\% outliers.

 \subsection{Review and Comparison}

We compare all our five models in Table \ref{compare}, and evaluate whether they bring in extra complexity\footnote{Here we refer to the parameter complexity. For example, the training data doubles  from Model II to III, but the amount of free model parameters keeps unchanged. Therefore, the parameter complexity is the same, but the training time definitely increases.}. \textbf{The general methodology to design and evolve models}, is to focus on adapting the model both to the specific task (SR pre-training, etc.), and to the specific data domains (LR-HR feature transfer, partially coupled, huber loss, etc.)

\section{Solving Real VLRR Problems}

We apply our proposed models to solve three VLRR applications. All model configurations, parameters and training details are similar to the CIFAR-10/100 simulation cases, unless otherwise specified.

\begin{figure*}[htbp]
\centering
\begin{minipage}{0.33\textwidth}
\centering \subfigure[] {
\includegraphics[width=\textwidth]{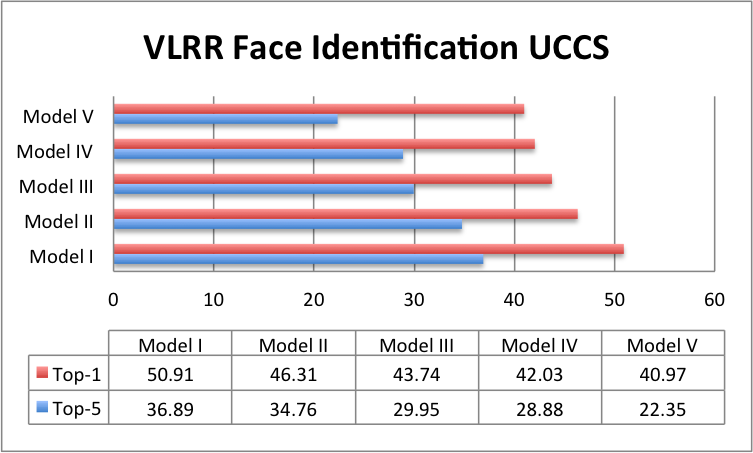}
}\end{minipage}
\begin{minipage}{0.33\textwidth}
\centering \subfigure[] {
\includegraphics[width=\textwidth]{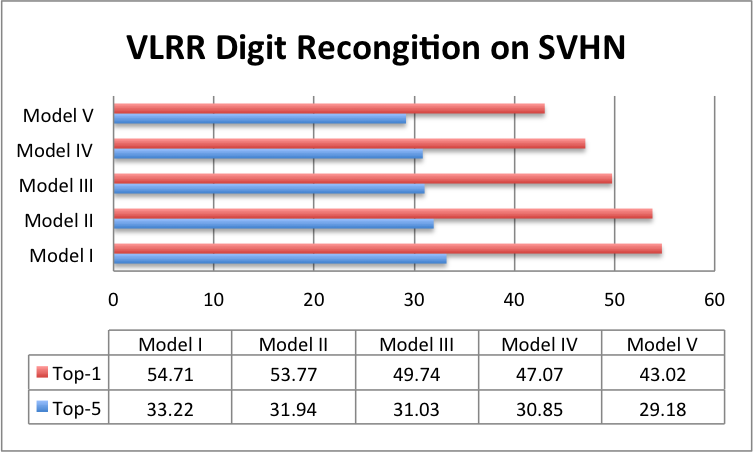}
}\end{minipage}
\begin{minipage}{0.33\textwidth}
\centering \subfigure [] {
\includegraphics[width=\textwidth]{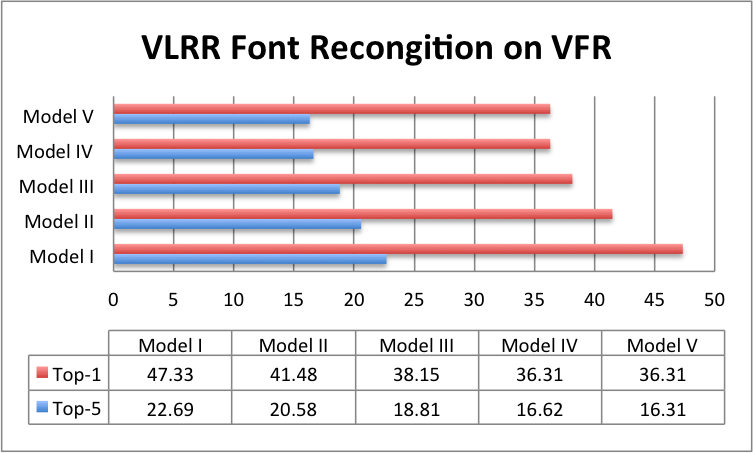}
}\end{minipage}
\caption{The top-1 and top-5 error rates of all five models, on three real VLRR applications.}
\label{real}
\end{figure*}

\subsection{VLRR Face Identification}

\noindent \textbf{Dataset} Although many face image datasets, such as LFW \cite{LFW}, SCface \cite{scface}, and CMU-PIE \cite{CMU}, have been developed, most of them were collected in controlled indoor environments with limited variability. A new UCCS dataset was recently proposed in \cite{UCCS} as a larger and more challenging benchmark for unconstrained face recognition. The authors of \cite{UCCS} reported a baseline face identification result of around 78\% top-1 accuracy on a 180-subject subset of original-resolution images, by extracting 73 face attributes \cite{kumar2009attribute} and using SVM for classification. We normalize the cropped face regions to 80 $\times$ 80 as HR, and downsample them by a factor $s$ = 5 for LR images of 16 $\times$ 16. We perform evaluations on a 180-subject subset, where each subject has 25 or more images. We use 4,500 images for training, and the remaining 935 for testing.
%

\noindent \textbf{Performance} Considering the training data is not sufficient (it is a common defect for unconstrained face datasets), we perform layer-by-layer greedy unsupervised training \cite{semi3} for all models\footnote{For Model I, the pre-training is only to reconstruct the LR image itself. For Model II to V, the pre-training is still SR-type, but is performed layer-by-layer instead of end-to-end.}.  As in Fig. \ref{real} (a), The performances consistently improve with our model evolves, finally reaching 40.97\%@top-1 and 22.35\%@top-5, with a significant margin of around 10\%@top-1 over the simplest baseline. Model V correctly classifies 552 out of 935 testing samples in top-1 results, and 726 in top-5. Model II reaches a 4.60\% improvement over Model I, while Model III further improves by 2.57\%. That identifies the key role of hallucinated features in this task, by both SR pre-training and LR-HR transfer. Besides, with more free model parameters, the partially-coupled architecture contributes to a performance rise of 1.71\%@top-1. Finally, the utility of huber loss brings an extra gain of 1.06\%@top-1.

\subsection{VLRR Digit Recognition}

\noindent \textbf{Dataset} SVHN \cite{SVH} is a real-world massive digit image dataset with highly complex scenes. Note the median of all digit height is 28 pixels, while more than 10,000 images have their original heights less than 10 pixels \cite{SVH}. That validates SVHN as a proper research object for VLRR. We utilize 604,388 (including the extra set) images for training and 26,032 for testing, covering 10 classes. All images are resized to 32 $\times$ 32 as HR, and LR images are obtained by downsampling with $s$ = 4 (8 $\times$ 8). For training, we have both HR and LR images available, while the testing set sees only LR images. Note that the preprocessing of SVHN introduces ``distracting digits'' to the sides of some digits of interest \cite{SVH}. Those strong outliers could largely confuse models, considering the highly noise-sensitive nature of very low resolution images \cite{VFR}.

\noindent \textbf{Performance} We calculate the top-1 and top-5 error rates for all five models. As in Fig. \ref{real} (b), Model still achieves the lowest error rates, with impressive records of 43.02\%@to-1 and 29.18\%@top-5. Note that the human top-1 accuracy, for original SVHN images of heights less than 25 pixels, is only around 82.0\% \cite{SVH}. Also, it is to our special attention that Model V obtains a 4.05\%@top-1 margin over Model IV, showing the great benefits of the robust huber loss, especially in the existence of strong outliers (distracting digits).  There is another performance gap of 4.03\%@top-1 between Model II and III, which proves that the LR-HR transfer enhances the feature learning. In addition, the fully-coupled and partially-coupled architecture demonstrate a 2.67\%@top-1 difference.




\subsection{VLRR Font Recognition}

\begin{figure}[thbp]
\centering
\begin{minipage}{0.40\textwidth}
\centering{
\includegraphics[width=\textwidth]{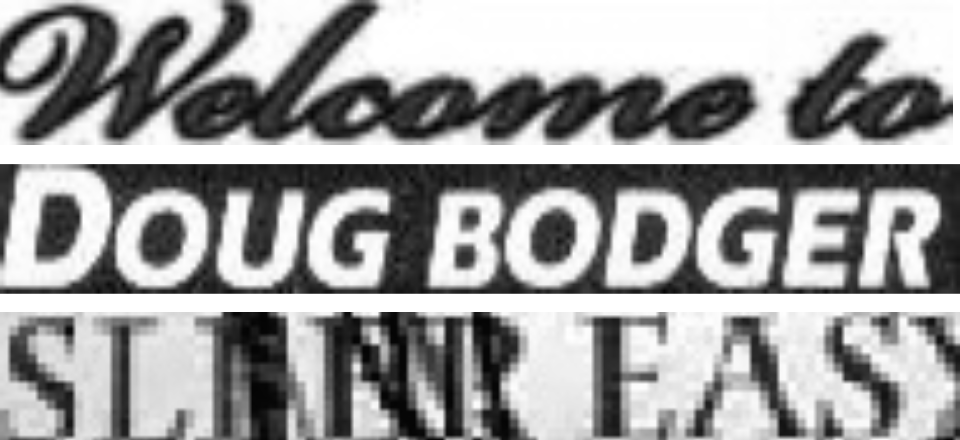}
}\end{minipage}
\caption{Examples (normalized) from the VFR testing set \cite{LFE}.}
\label{LFE}
 \vspace{-0.1em}
\end{figure}

\noindent \textbf{Dataset} The VFR dataset used in \cite{LFE, wang2015real, ICLR} includes 2,383 font classes. For each class, 1,000 synthetic text images were rendered for training, with a normalized height of 105.  A testing set was collected with 325 real-world images of 93 classes, and of various ROI (text) dimensions. Further statistics show that approximately 1/6 of the testing images (52) fall into VLRR cases (e.g, character heights are less than 16 pixels), and a majority of those ``VLRR images'' fail to be recognized even with the latest model \cite{ICLR}, due to the very heavy detail loss and compression artifacts after normalization. We aim to overcome it by explicitly taking resolution-enhancement into account during training, formulating a VLRR problem. All training images are downsampled by a \textbf{random} factor $s$ between [5, 15] to obtain LR images with heights between [7, 21] pixels, with the intention to make the learnt model robust to a wide range of very low resolutions in testing.

\noindent \textbf{Performance} With the identical training and testing sets, we fulfills the same task with an emphasis on improving recognition ability of VLRR subjects. We follow the data augmentations in \cite{ICLR} to train all models. While SR pre-training, LR-HR transfer and partially-coupled networks each still lead to visible gains, it is interesting to notice that Model V hardly benefits from the huber loss. That might be in part due to the very small sample size; but more importantly, the training set contains all rendered synthetic data without any outliers, which makes the robust loss unnecessary in unsupervised training. The proposed model reaches the error rates of 36.31\%@top-1 and 16.31\%@top-5. That outperforms the best accuracy numbers reported on VFR before, 38.15\%@top-1 and 20.62\%@top-5 \cite{ICLR}. A closer inspection reveals that our model correctly classify 33 out of 52  ``VLRR images'' as per top-5 results.

\section{Conclusion}

In this paper,  we study the challenging VLRR problem using the tool of deep networks. Starting from the simplest CNN baseline, we gradually evolve our models, with each step well motivated and justified. In addition to the large learning capacity, the final model also benefits from the SR pre-training,  the domain adaptation with partial flexibility, and the robust loss. The effectiveness of the proposed models is evaluated on three different VLRR tasks, with outstanding performances obtained.

{\small
\bibliographystyle{ieee}
\bibliography{egbib}
}

\end{document}